%% file: main.tex
\documentclass{article}
\pdfoutput=1

\usepackage[draft]{hyperref}
\usepackage{fancyref}
\usepackage{url}
\usepackage{microtype}
\usepackage{graphicx}
\usepackage{subfigure}
\usepackage{booktabs}       
\usepackage{amsfonts}       
\usepackage{nicefrac}       
\usepackage{color}
\usepackage{amsmath}
\usepackage{xcolor}
\usepackage{paralist}
\usepackage{enumerate}
\usepackage{paralist}
\usepackage{adjustbox}
\usepackage{algorithm,algorithmic}

\usepackage{threeparttable}
\usepackage{array}
\newcolumntype{C}[1]{>{\centering\let\newline\\\arraybackslash\hspace{0pt}}m{#1}}

\usepackage{caption} 
\usepackage[accepted]{icml2018}
\icmltitlerunning{Measuring abstract reasoning in neural networks}

\begin{document}

\twocolumn[
\icmltitle{Measuring abstract reasoning in neural networks}
\icmlsetsymbol{equal}{*}

\begin{icmlauthorlist}
\icmlauthor{David G.T. Barrett}{equal,goo}
\icmlauthor{Felix Hill}{equal,goo}
\icmlauthor{Adam Santoro}{equal,goo}
\icmlauthor{Ari S. Morcos}{goo}
\icmlauthor{Timothy Lillicrap}{goo}
\end{icmlauthorlist}

\icmlaffiliation{goo}{DeepMind, London, United Kingdom}
\icmlcorrespondingauthor{}{\{barrettdavid; felixhill; adamsantoro\}@google.com}
\icmlkeywords{Machine Learning, ICML}

\vskip 0.3in
]

\printAffiliationsAndNotice{*Equal contribution, ordered by surname.}  

\begin{abstract}
Whether neural networks can learn abstract reasoning or whether they merely rely on superficial statistics is a topic of recent debate. Here, we propose a dataset and challenge designed to probe abstract reasoning, inspired by a well-known human IQ test. To succeed at this challenge, models must cope with various generalisation `regimes' in which the training and test data differ in clearly-defined ways. We show that popular models such as ResNets perform poorly, even when the training and test sets differ only minimally, and we present a novel architecture, with a structure designed to encourage reasoning, that does significantly better. When we vary the way in which the test questions and training data differ, we find that our model is notably proficient at certain forms of generalisation, but notably weak at others. We further show that the model's ability to generalise improves markedly if it is trained to predict symbolic explanations for its answers. Altogether, we introduce and explore ways to both measure and induce stronger abstract reasoning in neural networks. Our freely-available dataset should motivate further progress in this direction.
\end{abstract}

\begingroup
\let\clearpage\relax
\include{intro}
\include{rpm}
\include{models}
\include{results}

\include{discussion}
\endgroup

\subsubsection*{Acknowledgments}
We would like to thank David Raposo, Daniel Zoran, Murray Shanahan, Sergio Gomez, Yee Whye Teh and Daan Wierstra for helpful discussions and all the DeepMind team for their support.
\bibliography{bibliography}
\bibliographystyle{icml2018}

\include{appendix}

\end{document}

%% file: intro.tex
\section{Introduction}

Abstract reasoning is a hallmark of human intelligence. A famous example is Einstein's elevator thought experiment, in which Einstein reasoned that an equivalence relation exists between an observer falling in uniform acceleration and an observer in a uniform gravitational field. It was the ability to relate these two abstract concepts that allowed him to derive the surprising predictions of general relativity, such as the curvature of space-time. 

\begin{figure}
    \centering
    \includegraphics[width=0.475\textwidth]{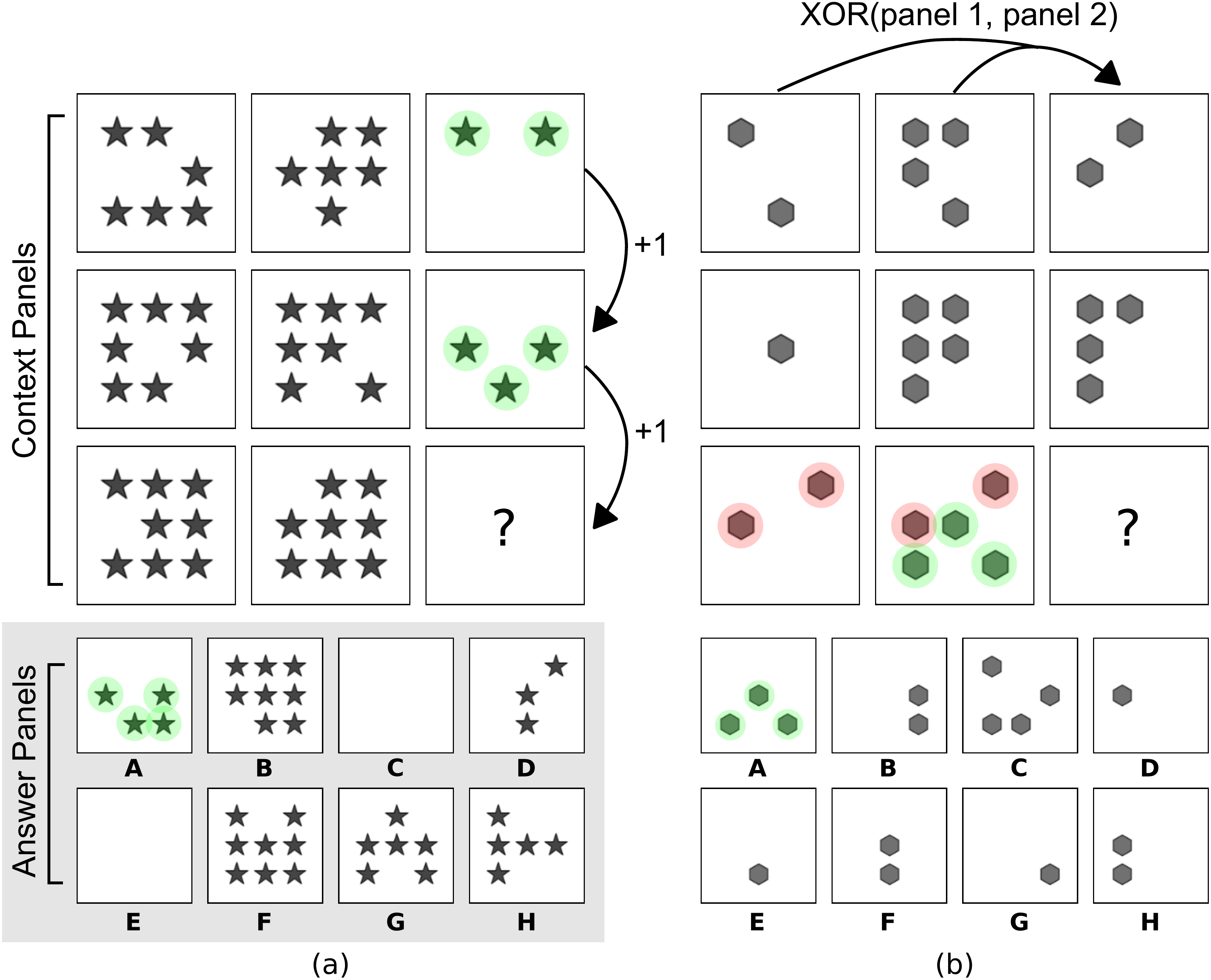}
    \caption{\textbf{Raven-style Progressive Matrices}. In (a) the underlying abstract rule is an arithmetic progression on the number of shapes along the columns. In (b) there is an \texttt{XOR} relation on the shape positions along the rows (panel 3 = \texttt{XOR}(panel 1, panel 2)). Other features such as shape type do not factor in. \textbf{A} is the correct choice for both.}
    \label{fig:rpm_intro}
\end{figure}

A human's capacity for abstract reasoning can be estimated surprisingly effectively using simple visual IQ tests, such as Raven's Progressive Matrices (RPMs) (Figure \ref{fig:rpm_intro}) \cite{raven1938raven}. The premise behind RPMs is simple: one must reason about the relationships between perceptually obvious visual features -- such as shape positions or line colors -- to choose an image that completes the matrix. For example, perhaps the size of squares increases along the rows, and the correct image is that which adheres to this size relation. RPMs are strongly diagnostic of abstract verbal, spatial and mathematical reasoning ability, discriminating even among populations of highly educated subjects~\cite{snow1984topography}. 

Since one of the goals of AI is to develop machines with similar abstract reasoning capabilities to humans, to aid scientific discovery for instance, it makes sense to ask whether visual IQ tests can help to understand learning machines. Unfortunately, even in the case of humans such tests can be invalidated if subjects prepare too much, since test-specific heuristics can be learned that shortcut the need for generally-applicable reasoning~\cite{te2001practice,flynn1987massive}. This potential pitfall is even more acute in the case of neural networks, given their striking capacity for memorization \cite{zhang2016understanding} and ability to exploit superficial statistical cues \cite{jo2017measuring, szegedy2013interesting}. 

Nonetheless, we contend that visual intelligence tests can help to better understand learning and reasoning in machines \citep{fleuret2011comparing}, provided they are coupled with a principled treatment of generalisation. Suppose we are concerned with whether a model can robustly infer the notion of `monotonically increasing'. In its most abstract form, this principle can apply to the quantity of shapes or lines, or even the intensity of their colour. We can construct training data that instantiates this notion for increasing quantities or sizes and we can construct test data that only involves increasing colour intensities. Generalisation to the test set would then be evidence of an abstract and flexible application of what it means to monotonically increase. In this way, a dataset with explicitly defined abstract semantics (e.g., relations, attributes, pixels, etc.), allows us to curate training and testing sets that precisely probe the generalisation dimensions of abstract reasoning in which we are interested.

To this end, we have developed a large dataset of abstract visual reasoning questions where the underlying abstract semantics can be precisely controlled. This approach allows us to address the following questions: (1) Can state-of-the-art neural networks find solutions -- \textit{any} solutions -- to complex, human-challenging abstract reasoning tasks if trained with plentiful training data? (2) If so, how well does this capacity generalise when the abstract content of training data is specifically controlled for? 

To begin, we describe and motivate our dataset, outline a procedure for automatic generation of data, and detail the generalisation regimes we chose to explore. Next, we establish a number of strong baselines, and show that well known architectures that use only convolutions, such as ResNet-50 \cite{he2016deep}, struggle. We designed a novel variant of the Relation Network \cite{santoro2017simple, raposo2017discovering}, a neural network with specific structure designed to encourage relation-level comparisons and reasoning. We found that this model substantially outperforms other well-known architectures. We then study this top-performing model on our proposed generalisation tests and find that it generalises well in certain test regimes (e.g. applying known abstract relationships in novel combinations), but fails notably in others (such as applying known abstract relationships to unfamiliar entities). Finally, we propose a means to improve generalisation: the use of auxiliary training to encourage our model to provide an explanation for its solutions.

%% file: rpm.tex
\section{Procedurally generating matrices}

In 1936 the psychologist John Raven introduced the now famous human IQ test: Raven's Progressive Matrices (RPM) \cite{raven1938raven}. RPMs consist of an incomplete $3 \times 3$ matrix of context images (see figure~\ref{fig:rpm_intro}), and some (typically 8) candidate answer images. The subject must decide which of the candidate images is the most appropriate choice to complete the matrix. 

It is thought that much of the power of RPMs as diagnostic of human intelligence derives from the way they probe \textit{eductive} or \textit{fluid} reasoning \cite{jaeggi2008improving}. Since no definition of an `appropriate'' choice is provided, it is in possible in principle to come up with a reason supporting any of the candidate answers. To succeed, however, the subject must assess all candidate answers, all plausible justifications for those answers, and identify the answer with the strongest justification. In practice, the right answer tends to be the one that can be explained with the simplest justification using the basic relations underlying the matrices. 

Although Raven hand-designed each of the matrices in his tests, later research typically employed some structured generative model to create large numbers of questions. In this setting, a potential answer is correct if it is consistent with the underlying generative model, and success rests on the ability to invert the model.

\begin{figure}
    \centering
    \includegraphics[width=0.475\textwidth]{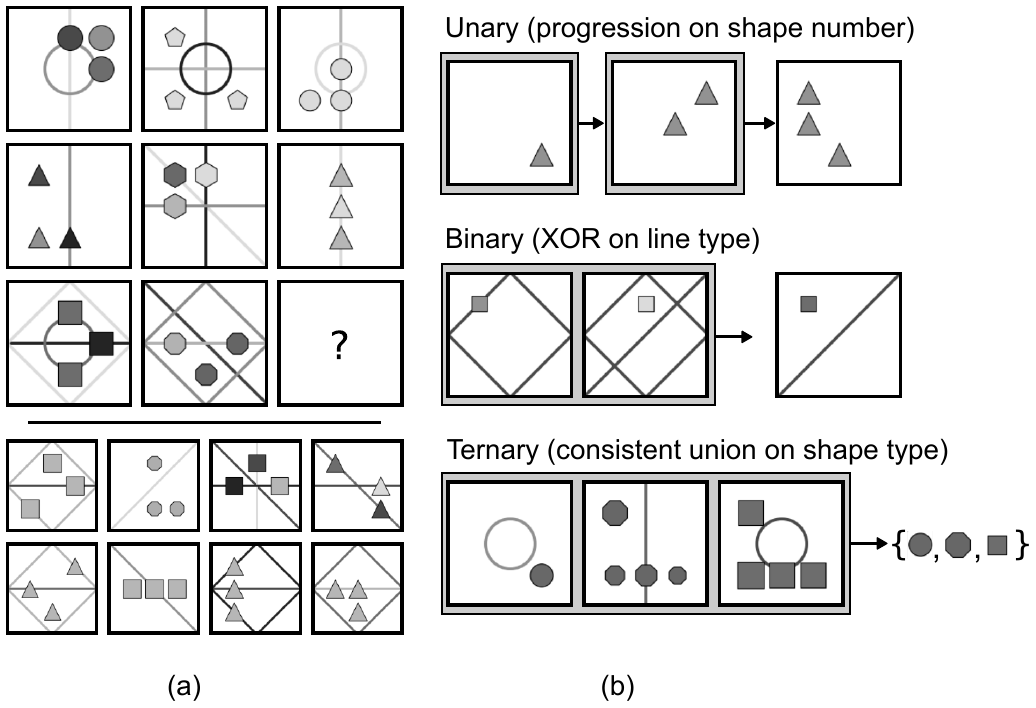}
    \caption{\textbf{A difficult PGM and a depiction of relation types}. (a) a challenging puzzle with multiple relations and distractor information. (b) a possible categorization of relation types based on how the panels are considered when computing the relation: for unary, a function is computed on one panel to produce the subsequent panel; for binary, two independently sampled panels are considered in conjunction to produce a third panel; and for ternary, all three panels adhere to some rule, such as all containing shapes from some common set, regardless of order.}
    \label{fig:rpm_hard}
\end{figure}

\subsection{Automatic generation of PGMs}
\label{sec:auto}

Here we describe our process for creating RPM-like matrices. We call our dataset the \textit{Procedurally Generated Matrices} (PGM) dataset. To generate PGMs, we take inspiration from \citet{carpenter1990one}, who identified and catalogued the relations that commonly underlie RPMs, as well as \citet{wang2015automatic}, who outlined one process for creating an automatic generator.

The first step is to build an abstract structure for the matrices. This is done by randomly sampling from the following primitive sets:
\begin{compactitem}
\item relation types ($\mathcal{R}$, with elements $r$): \texttt{progression}, \texttt{XOR}, \texttt{OR}, \texttt{AND}, \texttt{consistent union}\footnote{Consistent union is a relation wherein the three panels contain elements from some common set, e.g., shape types \{square, circle, triangle \}. The ordering of the panels containing the elements does not matter.}
\item object types ($\mathcal{O}$, with elements $o$): \texttt{shape}, \texttt{line}
\item attribute types ($\mathcal{A}$, with elements $a$): \texttt{size}, \texttt{type}, \texttt{colour}, \texttt{position}, \texttt{number}
\end{compactitem}

The structure $\mathcal{S}$ of a PGM is a set of triples, $\mathcal{S} = \{ [r, o, a] : r \in \mathcal{R}, o \in \mathcal{O}, a \in \mathcal{A} \}$. These triples determine the challenge posed by a particular matrix. For instance, if $\mathcal{S}$ contains the triple \texttt{[progression, shape, colour]}, the PGM will exhibit a progression relation, instantiated on the colour (greyscale intensity) of shapes. Challenging PGMs exhibit relations governed by multiple such triples: we permit up to four relations per matrix ($1\leq|\mathcal{S}|\leq4$). 

Each attribute type $a \in \mathcal{A}$ (e.g. \texttt{colour}) can take one of a finite number of discrete values $v \in \mathcal{V}$ (e.g. 10 integers between $[0, 255]$ denoting greyscale intensity). So a given structure has multiple realisations depending on the randomly chosen values for the attribute types, but all of these realisations share the same underlying abstract challenge. The choice of $r$ constrains the values of $v$ that can be realized. For instance, if $r$ is \texttt{progression}, the values of $v$ must strictly increase along rows or columns in the matrix, but can vary randomly within this constraint. See the appendix for the full list of relations, attribute types, values, their hierarchical organisation, and other statistics of the dataset.

We use $\mathcal{S}_a$ to denote the set of attributes among the triples in $\mathcal{S}$. After setting values for the \texttt{colour} attribute, we then choose values for all other attributes $a \not\in \mathcal{S}_a$ in one of two ways. In the \textit{distracting} setting, we allow these values to vary at random provided that they do not induce any further meaningful relations. Otherwise, the $a \not\in \mathcal{S}_a$ take a single value that remains consistent across the matrix (for example, perhaps all the shapes are the exact same size). Randomly varying values across the matrix is a type of distraction common to Raven's more difficult Progressive Matrices. 

Thus, the generation process consists of: (1) Sampling $1$-$4$ triples, (2) Sampling values $v \in \mathcal{V}$ for each $a \in \mathcal{S}_a$, adhering to the associated relation $r$, (3) Sampling values $v \in \mathcal{V}$ for each $a \not\in \mathcal{S}_a$, ensuring no spurious relation is induced, (4) Rendering the symbolic form into pixels. 

\subsection{Generalisation Regimes}
\label{sec:gen_regimes}
Generalisation in neural networks has been subject of lots of recent debate, with some emphasising the successes~\cite{lecun2015deep} and others the failures~\cite{garnelo2016towards,lake2017still,marcus2018deep}. Our choice of regimes is informed by this, but is in no way exhaustive. 

\paragraph{(1) Neutral}
In both training and test sets, the structures $S$ can contain any triples $[r, o, a]$ for $r \in \mathcal{R}$, $o \in \mathcal{O}$ and $a \in \mathcal{A}$. The training and test sets are disjoint, but this separation was at the level of the input variables (i.e., the pixel manifestations of the matrices).

\paragraph{(2) Interpolation; (3) Extrapolation}
As in the neutral split, $S$ consisted of any triples $[r, o, a]$. For interpolation, in the training set, when $a=$ \texttt{colour} or $a=$ \texttt{size} (the ordered attributes), the values of $a$ were restricted to even-indexed members of the discrete set $V_a$, whereas in the test set only odd-indexed values were permitted. For extrapolation, the values of $a$ were restricted to the lower half of their discrete set of values $V_a$ during training, whereas in the test set they took values in the upper half. Note that all $S$ contained some triple $[r, o, a]$ with $a=$ \texttt{colour} or $a=$ \texttt{size}. Thus, generalisation is required for every question in the test set.

\paragraph{(4) Held-out Attribute \texttt{shape-colour} or (5) \texttt{line-type}}
$\mathcal{S}$ in the training set contained no triples with $o=$ \texttt{shape} and $a=$ \texttt{colour}. All structures governing puzzles in the test set contained at least one triple with $o=$ \texttt{shape} and $a=$ \texttt{colour}. For comparison, we included a similar split in which triples were held-out if $o=$ \texttt{line} and $a=$ \texttt{type}. 

\paragraph{6: Held-out Triples}
In our dataset, there are 29 possible unique triples $[r,o,a]$. We allocated seven of these for the test set, at random, but such that each of the $a \in \mathcal{A}$ was represented exactly once in this set. These held-out triples never occurred in questions in the training set, and every $\mathcal{S}$ in the test set contained at least one of them. 

\paragraph{7: Held-out Pairs of Triples}
All $\mathcal{S}$ contained at least two triples, of which 400 are viable\footnote{Certain triples, such as \texttt{[progression, shape, number] and [progression, shape, XOR]} cannot occur together in the same PGM} $([r_1,o_1,a_1],[r_2,o_2,a_2]) = (t_1, t_2)$. We randomly allocated 360 to the training set and 40 to the test set. Members $(t_1, t_2)$ of the 40 held-out pairs did not occur together in structures $\mathcal{S}$ in the training set, and all structures $\mathcal{S}$ had at least one such pair $(t_1, t_2)$ as a subset.

\paragraph{8: Held-out Attribute Pairs}
$\mathcal{S}$ contained at least two triples. There are 20 (unordered) viable pairs of attributes $(a_1, a_2)$ such that for some $r_i, o_i$,  $([r_1,o_1,a_1],[r_2,o_2,a_2])$ is a viable triple pair.   $([r_1,o_1,a_1],[r_2,o_2,a_2]) = (t_1, t_2)$. We allocated 16 of these pairs for training and four for testing. For a pair $(a_1, a_2)$ in the test set, $\mathcal{S}$ in the training set contained triples with $a_1$ and $a_2$. In the test set, all $\mathcal{S}$ contained triples with $a_1$ and $a_2$. 

%% file: models.tex
\section{Models and Experimental Setup}

We first compared the performance of several standard deep neural networks on the neutral split of the PGM dataset. We also developed a novel architecture based on Relation Networks \cite{santoro2017simple}, that we call the Wild Relation Network (WReN), named in recognition of Mary Wild who contributed to the development of Raven's progressive matrices along with her husband John Raven.

\begin{figure*}
    \centering
    \includegraphics[width=0.975\textwidth]{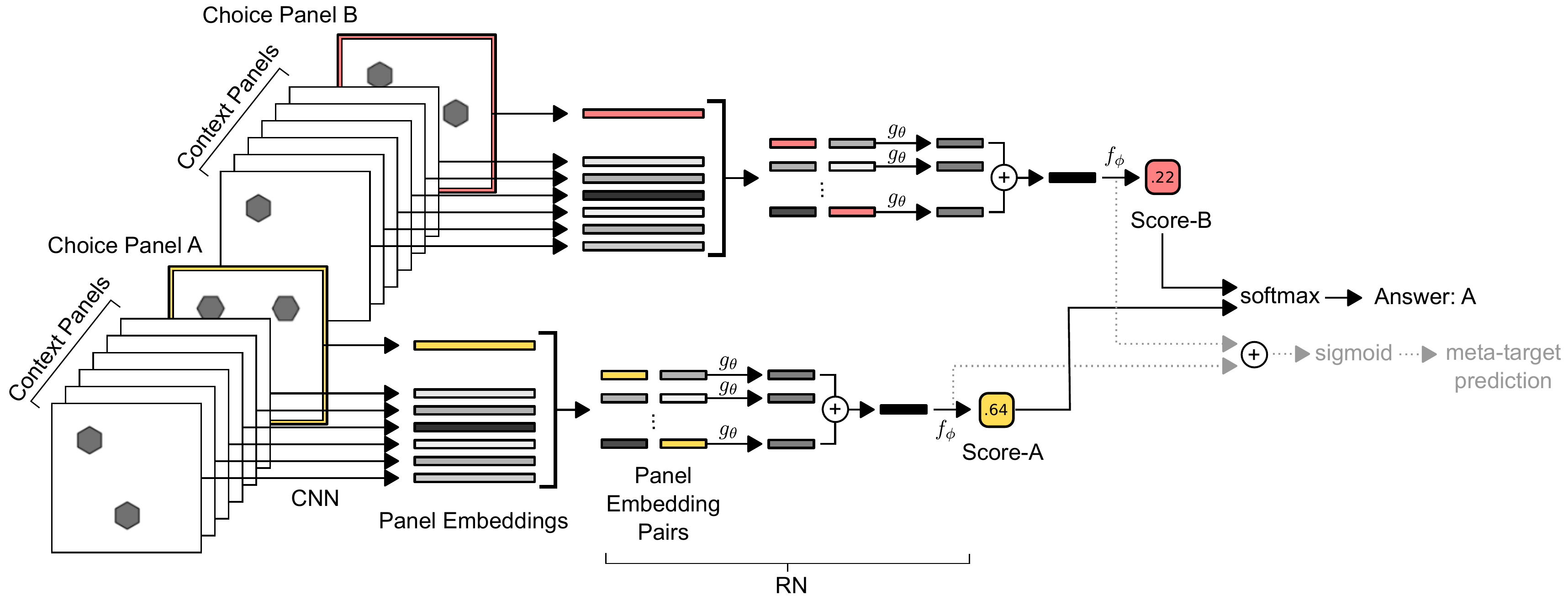}
    \caption{\textbf{WReN model} A CNN processes each context panel and an individual answer choice panel independently to produce $9$ vector embeddings. This set of embeddings is then passed to an RN, whose output is a single sigmoid unit encoding the ``score'' for the associated answer choice panel. $8$ such passes are made through this network (here we only depict $2$ for clarity), one for each answer choice, and the scores are put through a softmax function to determine the model's predicted answer.}
    \label{fig:wren}
\end{figure*}

The input consisted of the eight context panels and eight multiple-choice panels. Each panel is an $80 \times 80$ pixel image; so, the panels were presented as a set of $16$ feature maps.

Models were trained to produce the label of the correct missing panel as an output answer by optimising a softmax cross entropy loss. We trained all networks by stochastic gradient descent using the ADAM optimiser \cite{kingma2014adam}. For each model, hyper-parameters were chosen using a grid sweep to select the model with smallest loss estimated on a held-out validation set. We used the validation loss for early-stopping and we report performance values on a held-out test set. For hyper-parameter settings and further details on all models see appendix \ref{sec:Appendix}. 

\paragraph{CNN-MLP:} We implemented a standard four layer convolutional neural network with batch normalization and ReLU non-linearities \cite{lecun2015deep}. The set of PGM input panels was treated as a set of separate greyscale input feature maps for the CNN. The convolved output was passed through a two-layer, fully connected MLP using a ReLU non-linearity between linear layers and dropout of 0.5 on the penultimate layer. Note that this is the type of model applied to Raven-style sequential reasoning questions by~\citet{hoshen2017iq}. 

\paragraph{ResNet:} We used a standard implementation of the ResNet-50 architecture as described in \citet{he2016deep}. As before, each of the context panels and multiple-choice panels was treated as an input feature map. We also trained a selection of ResNet variants, including ResNet-101, ResNet-152, and several custom-built smaller ResNets. The best performing model was ResNet-50. 

\paragraph{LSTM:} We implemented a standard LSTM module \cite{Hochreiter1997}, based on \citet{zaremba2014recurrent}. Since LSTMs are designed to process inputs sequentially, we first passed each panel (context panels and multiple choice panels) sequentially and independently through a small $4$-layer CNN, tagged the CNN's output with a one-hot label indicating the panel's position (the top left PGM panel is tagged with label 1, the top-middle PGM panel is tagged with label 2 etc.), and passed the resulting sequence of labelled embeddings to the LSTM. The final hidden state of the LSTM was passed through a linear layer to produce logits for the softmax cross entropy loss. The network was trained using batch normalization after each convolutional layer and drop-out was applied to the LSTM hidden state.

\paragraph{Wild Relation Network (WReN):}
Our novel WReN model (fig. \ref{fig:wren}) applied a Relation Network module \cite{santoro2017simple} multiple times to infer the inter-panel relationships.

The model output a $1$-d score $s_k$ for a given candidate multiple-choice panel, with label $k\in [1,8]$. The choice with the highest score was selected as the answer $a$ using a softmax function $\sigma$ across all scores: $a = \sigma([s_1, \ldots, s_8])$. The score of a given multiple-choice panel was evaluated using a Relation Network (RN):
\begin{align}
\begin{split}\label{eq:1}
    s_k &= \text{RN}(\mathcal{X}_k)\\
        &= f_{\phi}\Big( \sum_{y, z \in \mathcal{X}_k } g_{\theta} (y, z)\Big),
\end{split}
\end{align}
where $\mathcal{X}_k = \left\{ x_1, x_2, ... ,x_8 \right\} \bigcup \left\{ c_k \right\} $, $c_k$ is the vector representation of the multiple choice panel $k$, and $x_i$ the representation of context panel $i$. The input vector representations were produced by processing each panel independently through a small CNN and tagging it with a panel label, similar to the LSTM processing described above, followed by a linear projection. The functions $f_{\phi}$ and $g_{\theta}$ are MLPs.

The structure of the WReN model is well matched to the problem of abstract reasoning, because it forms representations of pair-wise relations (using $g_{\theta}$), in this case, between each context panel and a given multiple choice candidate, and between context panels themselves. The function $f_{\phi}$ integrates information about context-context relations and context-multiple-choice relations to provide a score. Also the WReN model calculates a score for each multiple-choice candidate independently, allowing the network to exploit weight-sharing across multiple-choice candidates. 

\paragraph{Wild-ResNet:}
We also implemented a novel variant of the ResNet architecture in which one multiple-choice candidate panel, along with the eight context panels were provided as input, instead of providing all eight multiple-choices and eight context panels as input as in the standard ResNet. In this way, the Wild-ResNet is designed to provide a score for each candidate panel, independent of the other candidates. The candidate with the highest score is the output answer. This is similar to the WReN model described above, but using a ResNet instead of a Relation Network for computing a candidate score. 

\paragraph{Context-blind ResNet:}
A fully-blind model should be at chance performance level, which for the PGM task is $12.5\%$. However, sufficiently strong models can learn to exploit statistical regularities in multiple-choice problems using the choice inputs alone, without considering the context \cite{johnson2017clevr}. To understand the extent to which this was possible,  we trained a ResNet-50 model with only the eight multiple-choice panels as input. 

\subsection{Training on auxiliary information} 
\label{sec:aux}

We explored auxiliary training as a means to improve generalisation performance. We hypothesized that a model trained to predict the relevant relation, object and attribute types involved in each PGM might develop representations that were more amenable to generalisation. To test this, we constructed ``meta-targets'' encoding the relation, object and attribute types present in PGMs as a binary string. The strings were of length $12$, with elements following the syntax: (\texttt{shape, line, color, number, position, size, type, progression, XOR, OR, AND, consistent union}). We encoded each triple in this binary form, then performed an \texttt{OR} operation across all binary-encoded triple to produce the meta-target. That is, $\text{OR}([101000010000], [100100010000]) = [101100010000]$. The models then predicted these labels using a sigmoid unit for each element, trained with cross entropy. A scaling factor $\beta$ determined the influence of this loss relative to the loss computed for the answer panel targets: $\mathcal{L}_\text{total} = \mathcal{L}_\text{target} + \beta\mathcal{L}_\text{meta-target}$. We set $\beta$ to a non-zero value when we wish to explore the impact of auxiliary meta-target training.

%% file: results.tex
\clearpage

\section{Experiments}

\subsection{Comparing models on PGM questions}

We first compared all models on the Neutral train/test split, which corresponds most closely to traditional supervised learning regimes. Perhaps surprisingly given their effectiveness as powerful image processors, CNN models failed almost completely at PGM reasoning problems  (Table \ref{table:main}), achieving performance marginally better than our baseline - the context-blind ResNet model which is blind to the context and trained on only the eight candidate answers. The ability of the LSTM to consider individual candidate panels in sequence yielded a small improvement relative to the CNN. The best performing ResNet variant was ResNet-50, which outperformed the LSTM. ResNet-50 has significantly more convolutional layers than our simple CNN model, and hence has a greater capacity for reasoning about its input features.

The best performing model was the WReN model. This strong performance may be partly due to the Relation Network module, which was was designed explicitly for reasoning about the relations between objects, and partly due to the scoring structure. Note that the scoring structure is not sufficient to explain the improved performance as the WReN model substantially outperformed the best Wild-ResNet model, which also had a scoring structure.

\subsection{Performance on different question types}

Questions involving a single $[r, o, a]$ triple were easier than those involving multiple triples. Interestingly, PGMs with three triples proved more difficult than those with four. Although the problem is apparently more complex with four triples, there is also more available evidence for any solution. Among PGMs involving a single triple, \texttt{OR} ($64.7\%$) proved to be an easier relation than \texttt{XOR} ($53.2\%$). PGMs with structures involving lines ($78.3\%$) were easier than those involving shapes ($46.2\%$) and those involving \texttt{shape-number} were much easier ($80.1\%$) than those involving \texttt{shape-size} ($26.4\%$).This suggests that the model struggled to discern fine-grained differences in size compared to more salient changes such as the absence or presence of lines, or the quantity of shapes. For more details of performance by question type, see Appendix Tables \ref{table:RN_breakdown_number_of_relations}, \ref{table:RN_breakdown_relation_type}.

\subsection{Effect of distractors}

\begin{figure}
    \centering
    \includegraphics[width=0.45\textwidth]{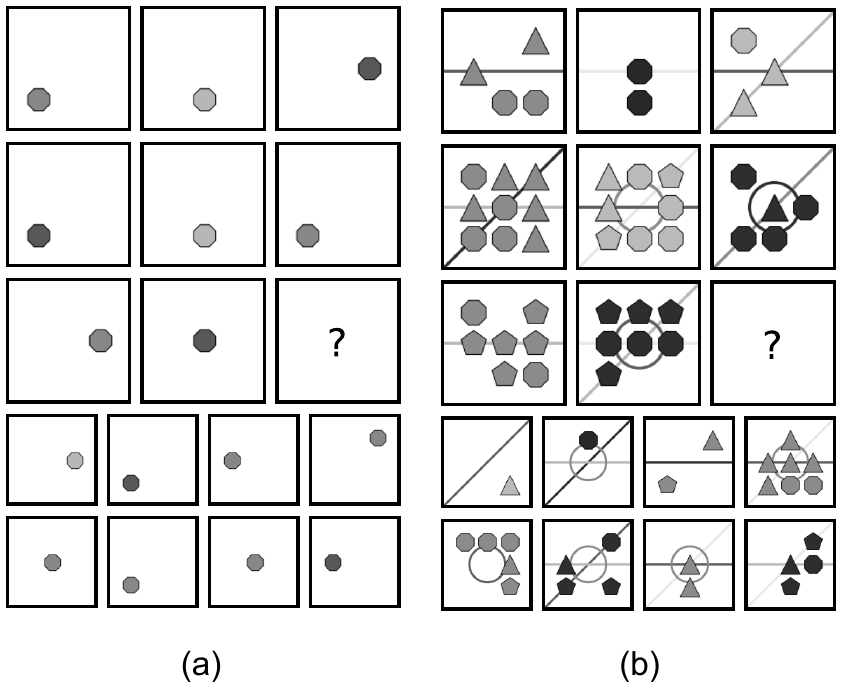}
    \caption{\textbf{The effect of distraction}. In both PGMs, the underlying structure $S$ is \texttt{[[shape, colour, consistent union]]}, but (b) includes distraction on \texttt{shape-number}, \texttt{shape-type}, \texttt{line-color}, and \texttt{line-type}.}
    \label{fig:distract}
\end{figure}

The results reported thus far were on questions that included distractor attribute values (see Fig. \ref{fig:distract}). The WReN model performed notably better when these distractors were removed ($79.3 \%$ on the validation and $78.3\%$ on the test set, compared with $63.0\%$ and $62.6\%$ with distractors). 

\subsection{Generalisation}

We compared the best performing WReN model on each of the generalisation regimes (Table \ref{table:main}), and observed notable differences in the ability of the model to generalise. Interpolation was the least problematic regime (generalisation error $14.6\%$). Note that performance on both the Interpolation and Extrapolation training sets was higher than on the neutral training set because certain attributes (\texttt{size}, \texttt{colour}) have half as many values in those cases, which reduces the complexity of the task.\footnote{Since test questions focus on held-out phenomena, test sets in different regimes may have differing underlying complexity. Absolute performance cannot therefore be compared across different regimes.} 

After Interpolation, the model generalised best in regimes where the test questions involved novel combinations of otherwise familiar $[r, o, a]$ triples (Held-out Attribute Pairs and Held-out Triple Pairs). This indicates that the model learned to combine relations and attributes, and did not simply memorize combinations of triples as distinct structures in their own right. However, worse generalisation in the case of Held-out Triples suggests that the model was less able to induce the meaning of unfamiliar triples from its knowledge of their constituent components. Moreover, it could not understand relations instantiated on entirely novel attributes (Held-out \texttt{line-type} , Held-out \texttt{shape-colour}). The worst generalisation was observed on the Extrapolation regime. Given that these questions have the same abstract semantic structure as interpolation questions, the failure to generalise may stem from the model's failure to perceive inputs outside of the range of its prior experience.

\begin{table*}[h]
\centering
\begin{threeparttable}
\begin{tabular}{c|ccc|ccc|ccc}
\multicolumn{3}{c}{} & \multicolumn{1}{c}{} & \multicolumn{3}{c}{$\beta=0$} & \multicolumn{3}{c}{$\beta=10$}\\
\textbf{Model} & \textbf{Test (\%)} && \textbf{Regime} & \textbf{Val. (\%)} & \textbf{Test (\%)} & \textbf{Diff.} & \textbf{Val. (\%)} & \textbf{Test (\%)} & \textbf{Diff.} \\
\cline{1-2}
\cline{4-10}
WReN& $\mathbf{62.6}$ && Neutral                     &  63.0     &  62.6     & \footnotesize{-0.6}   & 77.2 & 76.9   & \footnotesize{-0.3}  \\
Wild-ResNet & 48.0 && Interpolation               &  79.0     &  64.4     & \footnotesize{-14.6}  & 92.3 & 67.4   & \footnotesize{-24.9} \\
ResNet-50 & 42.0 && H.O. Attribute Pairs        &  46.7     &  27.2     & \footnotesize{-19.5}  & 73.4 & 51.7   & \footnotesize{-21.7} \\
LSTM & 35.8 && H.O. Triple Pairs           &  63.9     &  41.9     & \footnotesize{-22.0}  & 74.5 & 56.3   & \footnotesize{-18.2} \\
CNN + MLP & 33.0 && H.O. Triples                &  63.4     &  19.0     & \footnotesize{-44.4}  & 80.0 & 20.1   & \footnotesize{-59.9} \\
Blind ResNet & 22.4 && H.O. \texttt{line-type}     &  59.5     &  14.4     & \footnotesize{-45.1}  & 78.1 & 16.4   & \footnotesize{-61.7} \\
&&&H.O. \texttt{shape-colour}  &  59.1     &  12.5     & \footnotesize{-46.6}  & 85.2  & 13.0  & \footnotesize{-72.2} \\
&&&Extrapolation               &  69.3     &  17.2     & \footnotesize{-52.1}  & 93.6 & 15.5   & \footnotesize{-78.1} \\
\end{tabular}
\end{threeparttable}
\caption{Performance  of all models on the neutral split (left), and generalisation performance of the WReN model (right) with generalisation regimes ordered according to generalisation error for $\beta=0$. Context-blind ResNet generalisation test performances for all regimes is given in Table \ref{table:blind_generalisation_baselines} of the Appendix. (\textbf{Diff}: difference between test and validation performance, H.O:``Held-out")}  
\label{table:main}
\end{table*}

\subsection{Effect of auxiliary training}

We then explored the impact of auxiliary training on abstract reasoning and generalisation by training our models with symbolic meta targets as described in Section \ref{sec:aux}. In the neutral regime, we found that auxiliary training led to a $13.9\%$ improvement in test accuracy. Critically, this improvement in the overall ability of the model to capture the data also applied to other generalisation regimes. The difference was clearest in the cases where the model was required to recombine familiar triples into novel combinations: ($56.3\%$ accuracy on Held-out triple pairs, up from $41.9\%$, and $51.7\%$ accuracy on Held-out attribute pairs, up from $27.2\%$). Thus, the pressure to represent abstract semantic principles such that they can be decoded simply into discrete symbolic explanations seems to improve the ability of the model to productively compose its knowledge. This finding aligns with previous observations about the benefits of discrete channels for knowledge representation~\cite{andreas2016modular} and the benefit of inducing explanations or rationales~\cite{ling2017program}.  

\subsection{Analysis of auxiliary training}

In addition to improving performance, training with meta-targets provides a means to measure which shapes, attributes, and relations the model believes are present in a given PGM, providing insight into the model's decisions. Using these predictions, we asked how the WReN model's accuracy varied as a function of its meta-target predictions. Unsurprisingly, the WReN model achieved a test accuracy of 87.4\% when its meta-target predictions were correct, compared to only 34.8\% when its predictions were incorrect. 

The meta-target prediction can be broken down into predictions of object, attribute, and relation types. We leveraged these fine-grained predictions to ask how the WReN model's accuracy varied as a function of its predictions on each of these properties independently. The model accuracy increased somewhat when the shape meta-target prediction was correct ($78.2\%$) compared to being incorrect ($62.2\%$), and when attribute meta-target prediction was correct ($79.5\%$) compared to being incorrect ($49.0\%$). However, for the relation property, the difference between a correct and incorrect meta-target prediction was substantial ($86.8\%$ vs. $32.1\%$). This result suggests that predicting the relation property correctly is most critical to task success.

\begin{figure}
    \centering
    \includegraphics[width=0.35\textwidth]{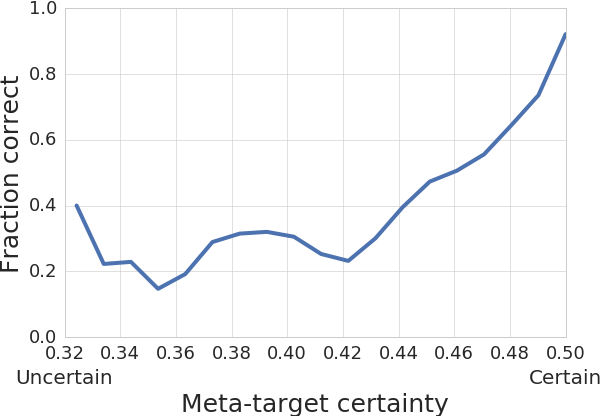}
    \caption{\textbf{Relationship between answer accuracy and meta-target prediction certainty for the WReN model ($\beta=10$)}. The WReN model is more accurate when it is more confident about its meta-target predictions. Certainty was defined as the mean absolute difference of the meta-target predictions from $0.5$.}
    \label{fig:meta_target_certainty}
\end{figure}

The model's prediction certainty, defined as the mean absolute difference of the meta-target predictions from $0.5$, was predictive of the model's performance, suggesting that the meta-target prediction certainty is an accurate measure of the model's confidence in an answer choice (Figure \ref{fig:meta_target_certainty}; qualitatively similar for sub-targets; Appendix Figures \ref{fig:meta_target_certainty_shape}-\ref{fig:meta_target_certainty_relation}).

%% file: discussion.tex
\section{Related work}
Various computational models for solving RPMs have been proposed in the cognitive science literature (see~\cite{lovett2017modeling} for a thorough review). The emphasis in these studies is on understanding the operations and comparisons commonly applied by humans. They typically factor out raw perception in favour of symbolic inputs, and hard-code strategies described by cognitive theories. In contrast, we consider models that process input from raw pixels and study how they infer, from knowledge of the correct answer, the processes and representations necessary to resolve the task. Much as we do, \citet{hoshen2017iq} trained neural networks to complete the rows or columns of Raven-style matrices from raw pixels. They found that a CNN-based model induced visual relations such as rotation or reflection, but they did not address the problem of resolving complete RPMs. Our experiments showed that such models perform poorly on full RPM questions. Moreover, \citet{hoshen2017iq} do not study generalisation to questions that differ substantively from their training data. \citet{wang2015automatic} present a method for automatically generating Raven-style matrices and verify their generator on humans, but do not attempt any modelling. Our method for automatically generating RPM-style questions borrowed extensively from the insights in that work.    

There is prior work emphasising both the advantages~\cite{clark2016my} and limitations~\cite{davis2014limitations} of applying standardized tests in AI (see \citet{marcus2016beyond} and contributed articles for a review). Approaches based on standardized testing generally focus on measuring the general knowledge of systems, while we focus on models' abilities to generalize learned information.

\section{Discussion}
One of the long-standing goals of artificial intelligence is to develop machines with abstract reasoning capabilities that equal or better those of humans. Though there has also been substantial progress in both reasoning and abstract representation learning in neural nets~\cite{botvinick2017building,lecun2015deep,higgins2016beta,higgins2017scan}, the extent to which these models exhibit anything like general abstract reasoning is the subject of much debate~\cite{garnelo2016towards,lake2017still,marcus2018deep}. The research presented here was therefore motivated by two main goals. (1) To understand whether, and (2) to understand how, deep neural networks might be able to solve abstract visual reasoning problems.

Our answer to (1) is that, with important caveats, neural networks can indeed learn to infer and apply abstract reasoning principles. Our best performing model learned to solve complex visual reasoning questions, and to do so, it needed to induce and detect from raw pixel input the presence of abstract notions such as logical operations and arithmetic progressions, and apply these principles to never-before observed stimuli. Importantly, we found that the architecture of the model made a critical difference to its ability to learn and execute such processes. While standard visual-processing models such as CNNs and ResNets performed poorly, a model that promoted the representation of, and comparison between parts of the stimuli performed very well. We found ways to improve this performance via additional supervision: the training outcomes and the model's ability to generalise were improved if it was required to decode its representations into symbols corresponding to the reason behind the correct answer.

When considering (2), it is important to note that our models were solving a very different problem from that solved by human subjects taking Raven-style IQ tests. The model's world was highly constrained, and its experience consisted of a small number of possible relations instantiated in finite sets of attributes and values across hundreds of thousands of examples. It is highly unlikely that the model's solutions match those applied by successful humans. This difference becomes clear when we study the ability of the model to generalise. Unlike humans, who must transfer knowledge distilled from their experience in everyday life to the unfamiliar setting of visual reasoning problems, our models exhibited transfer across question sets with a high degree of perceptual and structural uniformity. When required to interpolate between known attribute values, and also when applying known abstract content in unfamiliar combinations, the models generalised notably well. Even within this constrained domain, however, they performed strikingly poorly when required to extrapolate to inputs beyond their experience, or to deal with entirely unfamiliar attributes.

In this latter behaviour, the model differs in a crucial way from humans; a human that could apply a relation such as \texttt{XOR} to the colour of lines would almost certainly have no trouble applying it to the colour of shapes. On the other hand, even the human ability to extend apparently well-defined principles to novel objects has limits; this is precisely why RPMs are such an effective discriminator of human IQ. For instance, a human subject might be uncertain what it means to apply \texttt{XOR} to the size or shape of sets of objects, even if he or she had learned to do so perfectly in the case of colors.  

An important contribution of this work is the introduction of the PGM dataset, as a tool for studying both abstract reasoning and generalisation in models. Generalisation is a multi-faceted phenomenon; there is no single, objective way in which models can or should generalise beyond their experience. The PGM dataset provides a means to measure the generalization ability of models in different ways, each of which may be more or less interesting to researchers depending on their intended training setup and applications.

Designing and instantiating meaningful train/test distinctions to study generalisation in the PGM dataset was simplified by the objective semantics of the underlying generative model. Similar principles could be applied to more naturalistic data, particularly with crowdsourced human input. For instance, image processing models could be trained to identify black horses and tested on whether they can detect white horses, or trained to detect flying seagulls, flying sparrows and nesting seagulls, and tested on the detection of nesting sparrows. This approach was taken for one particular generalisation regime by \citet{ramakrishnan2017empirical}, who tested VQA models on images containing objects that were not observed in the training data. The PGM dataset extends and formalises this approach, with regimes that focus not only on how models could respond to novel factors or classes in the data, but also novel combinations of known factors etc.

In the next stage of this research, we will explore strategies for improving generalisation, such as meta-learning, and will further explore the use of richly structured, yet generally applicable, inductive biases. We also hope to develop a deeper understanding of the solutions learned by the WReN model when solving Raven-style matrices. Finally, we wish to end by inviting our colleagues across the machine learning community to participate in our new abstract reasoning challenge. 

%% file: appendix.tex
\clearpage

\appendix

\section{Appendix} 
\label{sec:Appendix}

\subsection{PGM Dataset} 
\label{sec:pgm_dataset}

Altogether there are 1.2M training set questions, 20K validation set questions, and 200K testing set questions.

When creating the matrices we aimed to use the full Cartesian product $\mathcal{R} \times \mathcal{A}$ for construction structures $\mathcal{S}$. However, some relation-attribute combinations are problematic, such as a progression on line type, and some attributes interact in interesting ways (such as number and position, which are in some sense tied), restricting the type of relations we can apply to these attributes. The final list of relevant relations per attribute type, broken down by object type (shape vs. line) is:

\textbf{shape}: \\
\-\hspace{1em} \textbf{size}: progression, XOR, OR, AND, consistent union\\
\-\hspace{1em} \textbf{color}: progression, XOR, OR, AND, consistent union\\
\-\hspace{1em} \textbf{number}: progression, consistent union\\
\-\hspace{1em} \textbf{position}: XOR, OR, AND\\
\-\hspace{1em} \textbf{type}: progression, XOR, OR, AND, consistent union\\
\textbf{line}:\\
\-\hspace{1em} \textbf{color}: progression, XOR, OR, AND, consistent union\\
\-\hspace{1em} \textbf{type}: XOR, OR, AND, consistent union\\

Since the number and position attribute types are tied (for example, having an arithmetic progression on number whilst having an XOR relation on position is not possible), we forbid number and position from co-occurring in the same matrix. Otherwise, all other $((r,o, a), (r, o, a))$ combinations occurred unless specifically controlled for in the generalisation regime.

We created a similar list for possible values for a given attribute:

\textbf{shape}: \\
\-\hspace{1em} \textbf{color}: 10 evenly spaced greyscale intensities in $[0, 1]$ \\
\-\hspace{1em} \textbf{size}: 10 scaling factors evenly spaced in $[0, 1]$ \footnote{The actual specific values used for size are numbers particular to the matplotlib implementation of the plots, and hence depend on the scale of the plot and axes, etc.}\\
\-\hspace{1em} \textbf{number}: 0, 1, 2, 3, 4, 5, 6, 7, 8, 9\\
\-\hspace{1em} \textbf{position} ((x, y) coordinates in a (0, 1) plot): \\
\-\hspace{2em}(0.25, 0.75), \\
\-\hspace{2em}(0.75, 0.75),\\
\-\hspace{2em}(0.75, 0.25),\\ 
\-\hspace{2em}(0.25, 0.25),\\ 
\-\hspace{2em}(0.5, 0.5),\\ 
\-\hspace{2em}(0.5, 0.25),\\ 
\-\hspace{2em}(0.5, 0.75),\\ 
\-\hspace{2em}(0.25, 0.5),\\ 
\-\hspace{2em}(0.75, 0.5)\\
\-\hspace{1em} \textbf{type}: circle, triangle, square, pentagon, hexagon, octagon, star\\

\textbf{line}:\\
\-\hspace{1em} \textbf{color}: 10 evenly spaced greyscale intensity in $[0, 1]$\\
\-\hspace{1em} \textbf{type}: diagonal down, diagonal up, vertical, horizontal, diamond, circle\\

\subsection{Examples of Raven-style PGMs} 
\label{sec:pgm_examples}
Given the radically different way in which visual reasoning tests are applied to humans (no prior experience) and to our models (controlled training and test splits), we believe it would be misleading to provide a human baseline for our results. However, for a sense of the difficulty of the task, we present here a set of 18 questions generated from the neutral splits. Note that the values are filtered for human readability. In the dataset there are 10 greyscale intensity values for shape and line colour and 10 sizes for each shape. In the following, we restrict to 4 clearly-distinct values for each of these attributes. Best viewed on a digital monitor, zoomed in (see next page). Informal human testing revealed wide variability: participants with a lot of experience with the tests could score well ($>80\%$), while others who came to the test blind would often fail to answer all the questions.

\begin{figure*}[h]
    \centering
    \includegraphics[width=0.8\textwidth]{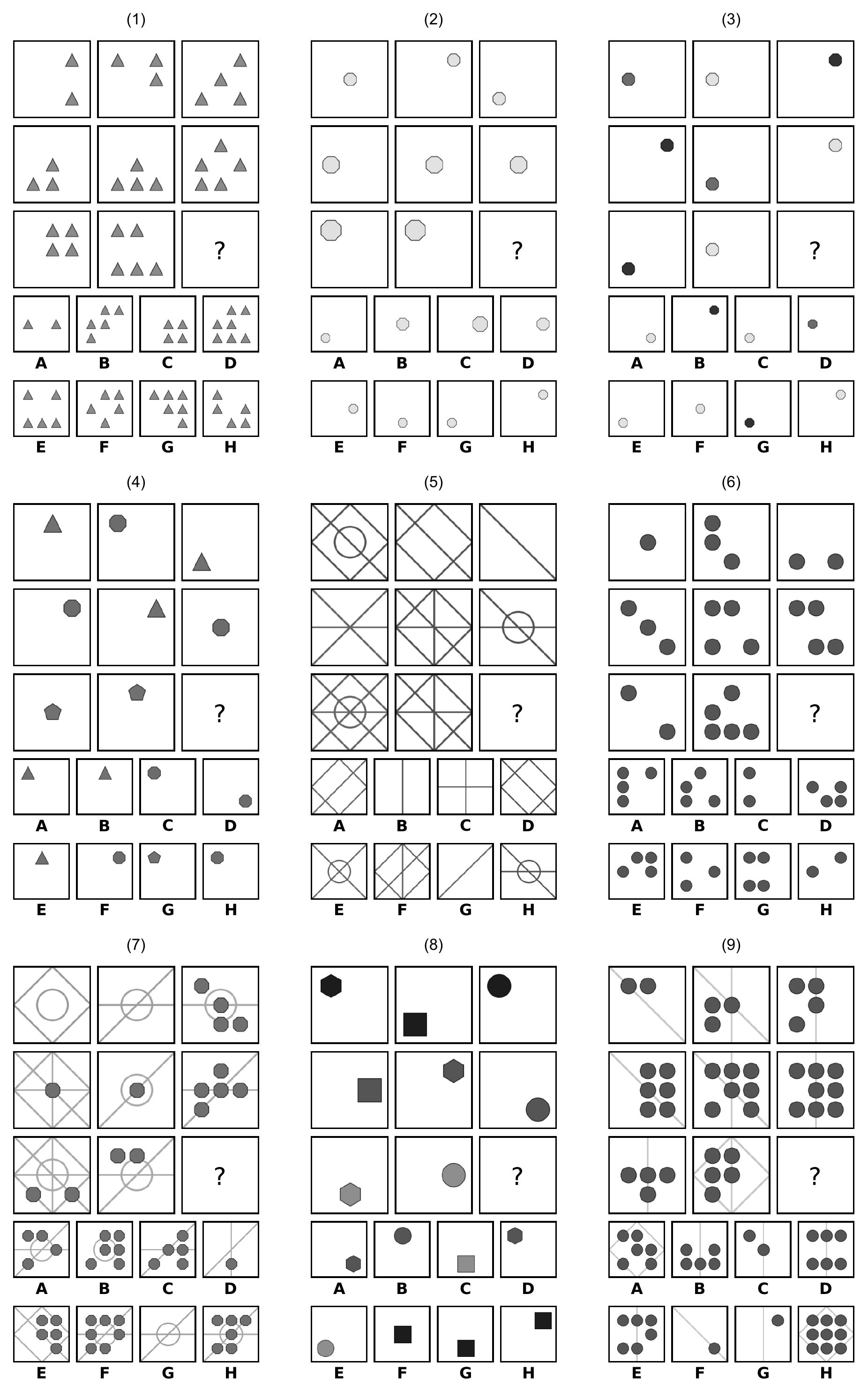}
\end{figure*}

\begin{figure*}[h]
    \centering
    \includegraphics[width=0.8\textwidth]{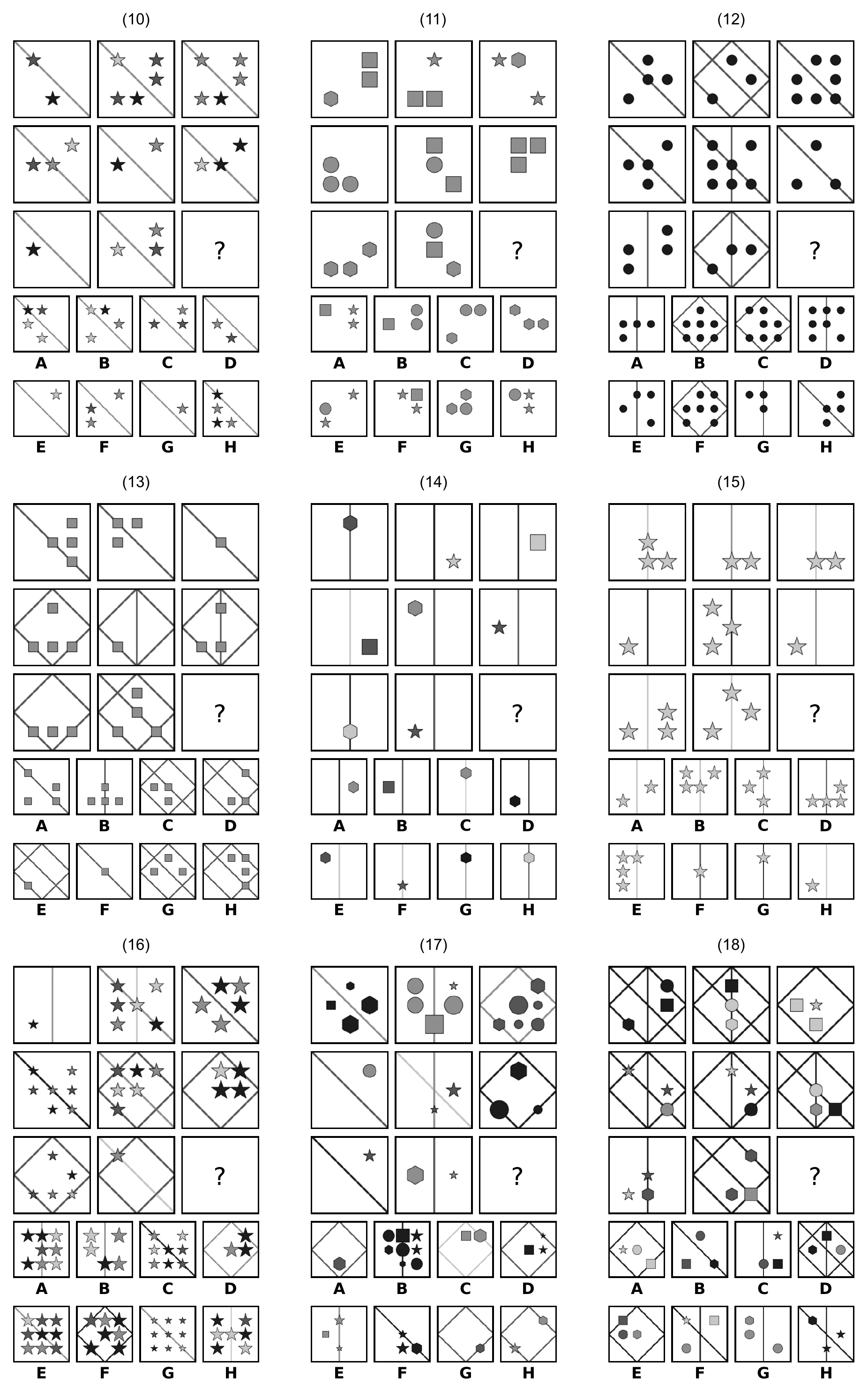}
\end{figure*}
\clearpage

\section{Model details} 
\label{sec:model_details}

Here we provide additional details for all our models, including the exact hyper-parameter settings that we considered. Throughout this section, we will use the notation [$x, y, z, w$] to describe CNN and MLP size. For a CNN, this notation refers to the number of kernels per layer: $x$ kernels in the first layer, $y$ kernels in the second layer, $z$ kernels in the third layer and w kernels in the fourth layer. For the MLP, it refers to the number of units per layer: $x$ units in the first layer, $y$ units in the second layer, $z$ units in the third layer and $w$ units in the fourth layer.

All models were trained using the Adam optimiser, with expoential decay rate parameters $\beta_1 = 0.9, \beta_2 = 0.999, \epsilon = 10^{-8}$. We also used a distributed training setup, using 4 GPU-workers per model.

\begin{table}[!h]
\centering
\begin{threeparttable}
\begin{tabular}{c|cc}
\toprule
 & hyper-parameters  \\
\hline
CNN kernels & [64, 64, 64, 64] \\
CNN kernel size & $3\times3$\\
CNN kernel stride & 2\\
MLP hidden-layer size & 1500\\
MLP drop-out fraction & 0.5 \\
Batch Size & 16  \\
Learning rate & 0.0003 \\
\bottomrule\addlinespace[1ex]
\end{tabular}
\end{threeparttable}
\caption{CNN-MLP hyper-parameters}
\label{table:CNN_hyper_parameters}
\end{table}

\begin{table}[!h]
\centering
\begin{threeparttable}
\begin{tabular}{c|cc}
\toprule
 & hyper-parameters  \\
\hline
Batch Size & 32  \\
Learning rate & 0.0003\\
\bottomrule\addlinespace[1ex]
\end{tabular}
\end{threeparttable}
\caption{ResNet-50 and context-blind ResNet hyper-parameters}
\label{table:Resnet_hyper_parameters}
\end{table}

\begin{table}[!h]
\centering
\begin{threeparttable}
\begin{tabular}{c|cc}
\toprule
 & hyper-parameters  \\
\hline
CNN kernels & [8, 8, 8, 8] \\
CNN kernel size & $3\times3$\\
CNN kernel stride & 2\\
LSTM hidden layer size & 96 \\
Drop-out fraction & 0.5 \\
Batch Size & 16  \\
Learning rate & 0.0001 \\
\bottomrule\addlinespace[1ex]
\end{tabular}
\end{threeparttable}
\caption{LSTM hyper-parameters}
\label{table:LSTM_hyper_parameters}
\end{table}

\begin{table}[!h]
\centering
\begin{threeparttable}
\begin{tabular}{c|cc}
\toprule
 & hyper-parameters  \\
\hline
CNN kernels & [32, 32, 32, 32] \\
CNN kernel size & $3\times3$\\
CNN kernel stride & 2\\
RN embedding size  & 256 \\
RN $g_\theta$ MLP & [512, 512, 512, 512] \\
RN $f_\phi$ MLP & [256, 256, 13] \\
Drop-out fraction & 0.5 \\
Batch Size & 32  \\
Learning rate & 0.0001 \\
\bottomrule\addlinespace[1ex]
\end{tabular}
\end{threeparttable}
\caption{WReN hyper-parameters}
\label{table:RN_scoring_hyper_parameters}
\end{table}

\begin{table}[!h]
\centering
\begin{threeparttable}
\begin{tabular}{c|cc}
\toprule
 & hyper-parameters  \\
\hline
Batch Size & 16  \\
Learning rate & 0.0003\\
\bottomrule\addlinespace[1ex]
\end{tabular}
\end{threeparttable}
\caption{Wild-ResNet hyper-parameters}
\label{table:Wild_Resnet_hyper_parameters}
\end{table}

\newpage

\section*{}
\clearpage

\section{Results}

\begin{table}[!h]
\centering
\begin{threeparttable}
\begin{tabular}{c||c|c}
\toprule
$\#$ Relations & WReN ($\%$) & Blind ($\%$)
\\
\hline
One & 68.5 & 23.6 \\
Two & 51.1 & 21.2\\
Three & 44.5 & 22.1\\
Four & 48.4 &  23.5 \\
\hline
All & 62.6 & 22.8\\
\bottomrule\addlinespace[1ex]
\end{tabular}
\end{threeparttable}
\caption{WReN test performance and Context-Blind ResNet performance after training on the neutral PGM dataset, broken down according to the number of relations per matrix.}
\label{table:RN_breakdown_number_of_relations}
\end{table}

\begin{table}[!h]
\centering
\begin{threeparttable}
\begin{tabular}{c||c|c}
\toprule& WReN ($\%$) & Blind ($\%$) \\
\hline
OR & 64.7 & 30.1\\
AND & 63.2 & 17.2 \\
consistent union & 60.1 & 28.0\\
progression  & 55.4 & 15.7\\
XOR & 53.2 & 20.2\\
\hline
number & 80.1 & 18.1 \\
position & 77.3 & 27.5 \\
type & 61.0 & 28.1 \\
color & 58.9 & 18.7 \\
size & 26.4 & 16.3 \\
\hline
line & 78.3 & 27.5 \\
shape & 46.2 & 18.6\\
\hline
All Single Relations & 68.5 & 23.6\\
\bottomrule\addlinespace[1ex]
\end{tabular}
\end{threeparttable}
\caption{WReN test performance and Context-Blind ResNet performance for single-relation PGM questions after training on the neutral PGM dataset, broken down according to the relation type, attribute type and object type in a given matrix.}
\label{table:RN_breakdown_relation_type}
\end{table}

\begin{table}[!t]
\centering
\begin{threeparttable}
\begin{tabular}{c||c|c}
\multicolumn{1}{c}{} &  \multicolumn{2}{c}{\textbf{Test (\%)} } \\
\toprule
\textbf{Regime} & $\mathbf{\beta=0}$   & $\mathbf{\beta=10}$  \\
\hline
Neutral                     &  22.4  & 13.5 \\
Interpolation               & 18.4   & 12.2 \\
H.O. Attribute Pairs       & 12.7 & 12.3 \\
H.O. Triple Pairs            & 15.0   & 12.6 \\
H.O. Triples                &  11.6  & 12.4\\
H.O. \texttt{line-type}     &   14.4 & 12.6\\
H.O. \texttt{shape-colour}  &  12.5  & 12.3 \\
Extrapolation               & 14.1   & 13.0 \\
\bottomrule\addlinespace[1ex]
\end{tabular}
\end{threeparttable}
\caption{Performance of the Context-blind Resnet model for all the generalization regimes, in the case where there is an additional auxiliary meta-target ($\beta=10$) and  in the case where there is no auxiliary meta-target  ($\beta=0$). Note that most of these values are either close to chance or slightly above chance, indicating that this baseline model struggles to learn solutions that generalise better than a random guessing solution. For several generalisation regimes such as Interplolation, H.O Attribute Pairs, H.O. Triples and H.O Triple Pairs the generalisation performance of the WReN model reported in Table \ref{table:main} is far greater than the generalisation performance of our context-blind baseline, indicating that the WReN generalisation cannot be accounted for with a context-blind solution.} 
\label{table:blind_generalisation_baselines}
\end{table}

\begin{figure}
    \centering
    \includegraphics[width=0.35\textwidth]{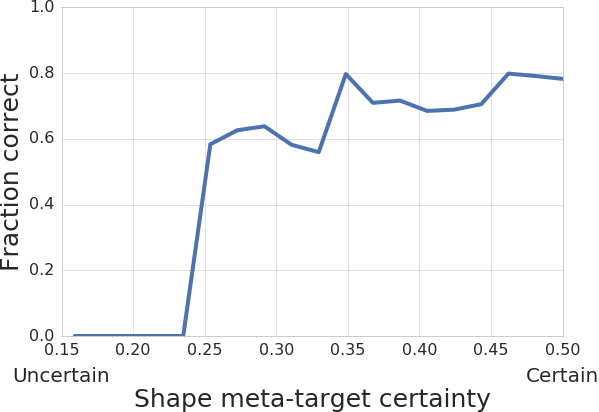}
    \caption{\textbf{Relationship between answer accuracy and shape meta-target prediction certainty}. The WReN model ($\beta=10$) is more accurate when confident about its meta-target predictions. Certainty was defined as the mean absolute difference of the predictions from $0.5$.}
    \label{fig:meta_target_certainty_shape}
\end{figure}

\begin{figure}
    \centering
    \includegraphics[width=0.35\textwidth]{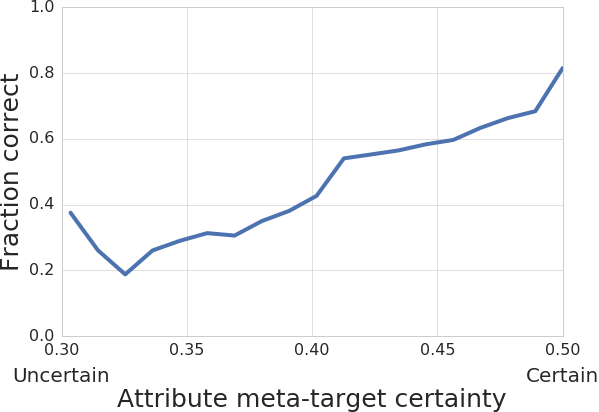}
    \caption{\textbf{Relationship between answer accuracy and attribute meta-target prediction certainty}}
    \label{fig:meta_target_certainty_attribute}
\end{figure}

\begin{figure}
    \centering
    \includegraphics[width=0.35\textwidth]{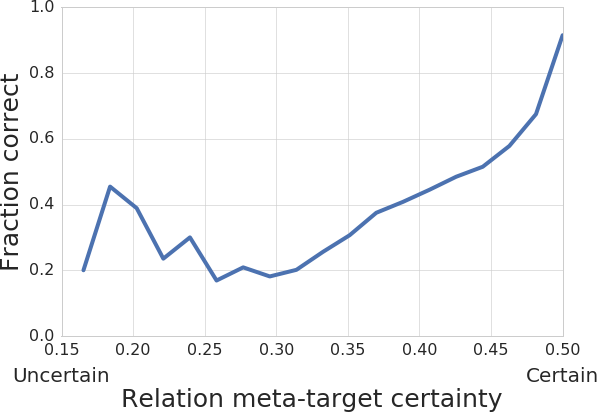}
    \caption{\textbf{Relationship between answer accuracy and relation meta-target prediction certainty}}
    \label{fig:meta_target_certainty_relation}
\end{figure}

\clearpage

\begin{figure*}
    \centering
    \includegraphics[width=0.75\textwidth]{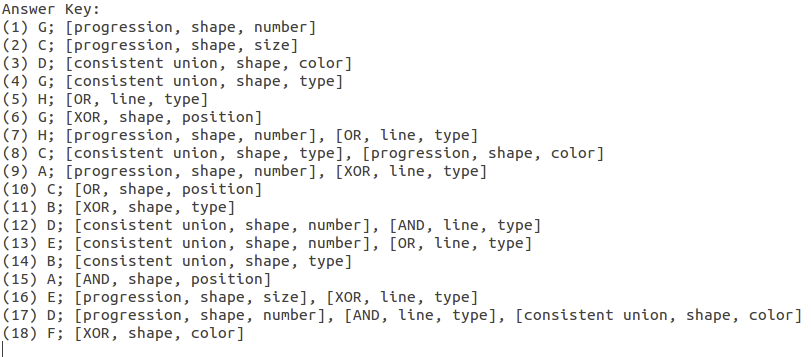}
    \caption{Answer key to puzzles in section A.2}
\end{figure*}